\def\year{2020}\relax
\documentclass[letterpaper]{article} 
\usepackage{aaai20}  
\usepackage{times}  
\usepackage{helvet} 
\usepackage{courier}  
\usepackage[hyphens]{url}  
\usepackage{graphicx} 
\usepackage{graphicx}  
\usepackage{algorithm}
\usepackage{multirow}
\usepackage{amsfonts}
\usepackage{amsmath}
\title{Learning Meta Model for Zero- and Few-shot Face Anti-spoofing}

\author{Yunxiao Qin,\textsuperscript{\rm 1,2} 
	Chenxu Zhao,\textsuperscript{\rm 2}\thanks{Corresponding Author.} 
	Xiangyu Zhu,\textsuperscript{\rm 3}
	Zezheng Wang,\textsuperscript{\rm 2} 
	Zitong Yu,\textsuperscript{\rm 4} \\ 
	\bf \Large Tianyu Fu,\textsuperscript{\rm 5} 
	Feng Zhou,\textsuperscript{\rm 2} 
	Jingping Shi,\textsuperscript{\rm 1} 
	Zhen Lei\textsuperscript{\rm 3} \\ 
	\textsuperscript{\rm 1}Northwestern Polytechnical University, Xian, China, 
	\textsuperscript{\rm 2}AIBEE, Beijing, China \\
	\textsuperscript{\rm 3}National Laboratory of Pattern Recognition, Institute of Automation, Chinese Academy of Science, Beijing, China \\
	\textsuperscript{\rm 4}CMVS, University of Oulu, Oulu, Finland, 
	\textsuperscript{\rm 5}Winsense Technology Ltd, Beijing, China \\
	qyxqyx@mail.nwpu.edu.cn, 
	\{cxzhao; zezhengwang; fzhou\}@aibee.com, 
	\{xiangyu.zhu; zlei\}@nlpr.ia.ac.cn \\
	zitong.yu@oulu.fi, 
	futianyu514@gmail.com, 
	shijingping@nwpu.edu.cn
}

\begin{document}
	\maketitle
	
	\begin{abstract}
		Face anti-spoofing is crucial to the security of face recognition systems. Most previous methods formulate face anti-spoofing as a supervised learning problem to detect various predefined presentation attacks, which need large scale training data to cover as many attacks as possible. 
		However, the trained model is easy to overfit several common attacks and is still vulnerable to unseen attacks.
		To overcome this challenge, the detector should: 1) learn discriminative features that can generalize to unseen spoofing types from predefined presentation attacks; 2) quickly adapt to new spoofing types by learning from both the predefined attacks and a few examples of the new spoofing types. 
		Therefore, we define face anti-spoofing as a zero- and few-shot learning problem.
		In this paper, we propose a novel \textbf{A}daptive \textbf{I}nner-update \textbf{M}eta \textbf{F}ace \textbf{A}nti-\textbf{S}poofing (AIM-FAS) method to tackle this problem through meta-learning. 
		Specifically, AIM-FAS trains a meta-learner focusing on the task of detecting unseen spoofing types by learning from predefined living and spoofing faces and a few examples of new attacks.
		To assess the proposed approach, we propose several benchmarks for zero- and few-shot FAS.
		Experiments show its superior performances on the presented benchmarks to existing methods in existing zero-shot FAS protocols.
	\end{abstract}
	
	\section{Introduction}
	
	Face recognition is a ubiquitous technology used in industrial applications and commercial products.
	However, face recognition system is easy to be fooled by presentation attacks (PAs),
	such as printed face (print attack), face replay on digital device (replay attack), face covered by a mask (3D-mask attack), etc.
	As a result, face anti-spoofing (FAS) system, which detects whether the presented face is live or not, becomes essential to keep the recognition system safe.
	
	Until now, researchers have proposed lots of hand-crafted feature based~\cite{Boulkenafet2017Face,Gan20173D,Lucena2017Transfer} 
	and deep-learning based methods\cite{Lucena2017Transfer,Xu2016Learning,Shao2017Deep} to discriminate spoof faces from living faces.
	Most of them train the detector to learn how to discriminate living and spoofing faces with numerous predefined living and spoofing faces in a supervised way.
	
	\begin{figure}[t]
		\centering
		\includegraphics[width=0.9\columnwidth]{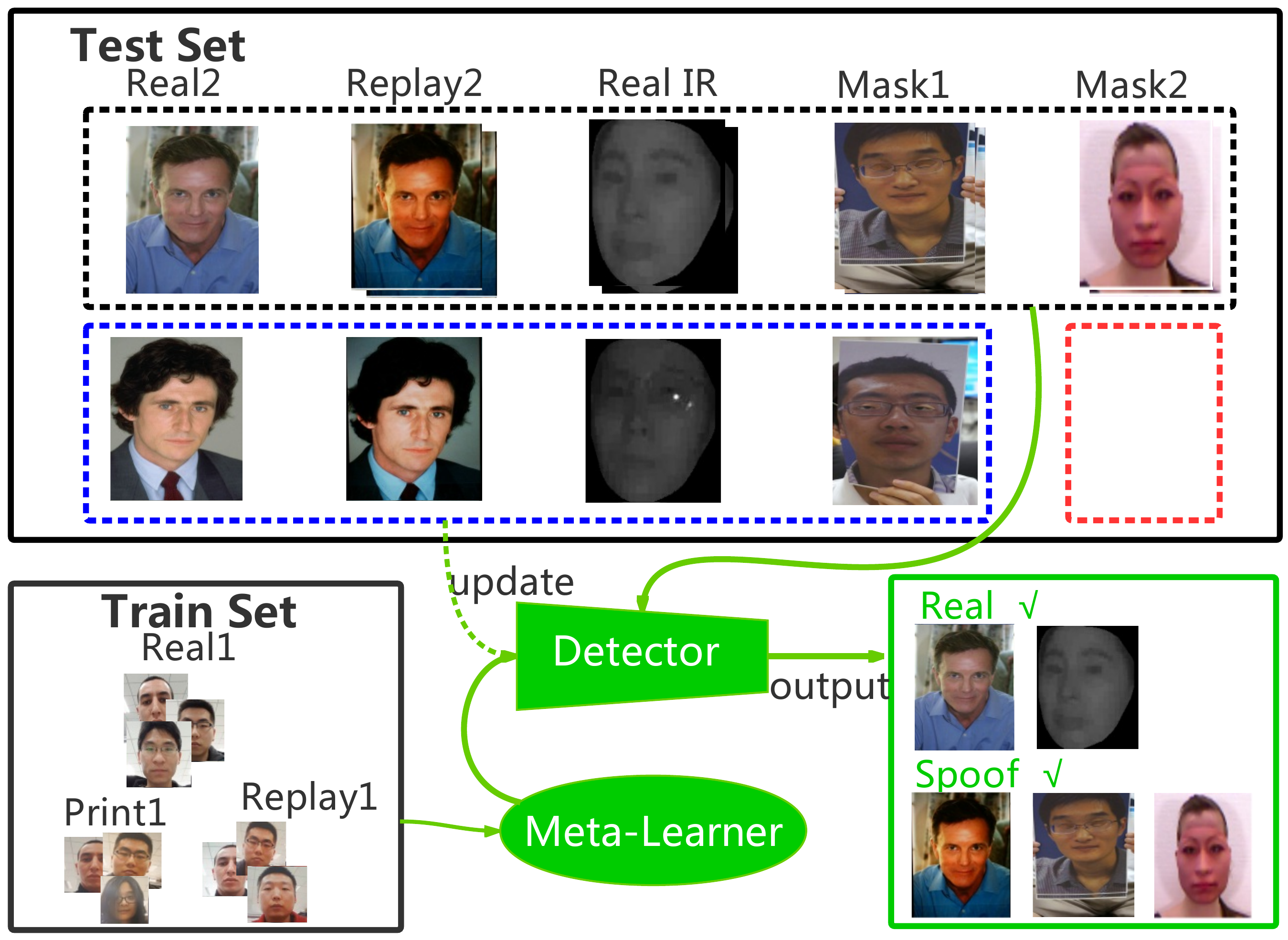}
		\caption{
			A zero- and few-shot FAS example.
			The train set contains several predefined living and spoofing types.
			The test set contains several faces of new emerged living and spoofing types.
			Zero-shot FAS is training the detector only on the train set and testing it on the test set.
			Whereas, few-shot FAS utilizes both the train set and a few collected faces (the blue box) for updating the detector.
		}
		\label{fig:fig1}
	\end{figure}

	The detectors are satisfactory at detecting the predefined PAs due to their data-driven training manner.
	However, when deployed in real scenarios, the FAS systems will encounter the following practical challenges. 
	\begin{itemize}
		\item A variety of application scenarios and unpredictable novel PAs keep evolving.
		Data-driven models may give unpredictable results when faced with out-of-distribution living examples captured in new application scenarios and spoofing examples with new PAs.
		\item When we adapt the anti-spoofing model to new attacks, existing methods need to collect sufficient samples for training. However, it is expensive to collect labeled data for every new attack since the spoofing keeps evolving. 
	\end{itemize}

	To overcome these challenges, we propose that FAS should be treated as an open set zero- and few-shot learning problem. As shown in Fig.1,
	\textit{Zero-shot learning} aims to learn general discriminative features, which are effective to detect un-predicted new PAs, from the predefined PAs. \textit{Few-shot learning} aims to quickly adapt the anti-spoofing model to new attacks by learning from both the predefined PAs and the collected very few examples of the new attack.
	
	Zero-shot FAS problem has been studied in \cite{DTN,zero-protocol-FAS} neglecting the few-shot scene.
	As aforementioned, the FAS detector should solve both zero- and few-shot FAS problems.
	To this end, inspired by the model-agnostic meta-learning (MAML)~\cite{MAML},
	we propose a novel meta-learning based FAS method: Adaptive Inner-update Meta Face Anti-spoofing (AIM-FAS).
	
	AIM-FAS solves the zero- and few-shot FAS problem by \textbf{F}usion \textbf{T}raining (FT) a meta-learner on zero- and few-shot FAS tasks, with \textbf{A}daptive \textbf{I}nner-\textbf{U}pdate (AIU) learning rate strategy.
	FT means the meta-learner is forced to focus on simultaneously learning:
	1) general discriminative features to detect unseen PAs from predefined PAs, if no instance of the new PA has been collected; 
	2) better discriminative features to adapt to new PA from both the predefined PAs and the few instances of the new PA, once a few instances of the new PA are collected.
	AIU means the meta-learner inner-updates with a learn-able regular inner-update step size.
	
	To evaluate the zero- and few-shot FAS, we propose three benchmarks to assess the FAS model's learning capability of detecting new PAs from the same domain, different domains, and different modals.
	
	The main contributions of this paper are:
	\begin{itemize}
		\item To the best of our knowledge, we are the first to formulate FAS as a zero- and few-shot learning problem.
		\item To solve zero- and few-shot FAS problem, we propose a novel meta-learning based approach: Adaptive Inner-update Meta Face Anti-spoofing (AIM-FAS), which Fusion Trains (FT) a meta-learner on zero- and few-shot FAS tasks with a novel developed Adaptive Inner-Update (AIU) strategy.
		\item We propose three novel zero- and few-shot FAS benchmarks to validate the efficacy of AIM-FAS. 
		\item Comprehensive experiments are conducted to show that AIM-FAS achieves state-of-the-art results on zero- and few-shot anti-spoofing benchmarks. 
	\end{itemize}

	\section{Background}
	
	\subsection{Face Anti-Spoofing}
	
	Traditional FAS methods~\cite{Pereira2012LBP,Pereira2013Can,Maatta2011Face,Patel2016Secure,Boulkenafet2017Face_SURF,Komulainen2014Context} usually extract hand-crafted features from the facial images and train a binary classifier to detect spoofing faces. 
	Recently, deep learning based FAS methods~\cite{Lucena2017Transfer,Nagpal2018A,Li2017An,Patel2016Cross} attract more attention.  
	These methods commonly train a deep network to learn static discrimination between living and spoofing faces, with binary classification or depth regression supervision.
	Recent researches show that the depth regression supervised methods~\cite{Atoum2018Face,Liu2018Learning} outperform the binary classification methods, mainly because they provide the network with more detail information to study the spoofing cues.
	However, either traditional or deep learning based approaches are still sensitive to various conditions, such as illumination, blur pattern, capture camera, and presentation attack instruments.
	Slight change of these conditions would significantly affect the performance of the FAS detector.

	\begin{figure*}[t]
		\centering
		\includegraphics[width=2\columnwidth]{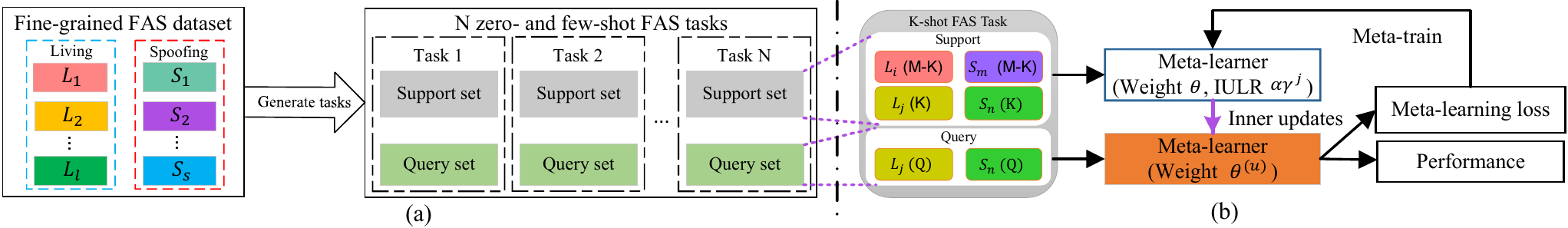}
		\caption{
			(a) The fine-grained FAS dataset contains several living ($L_1, L_2, ..., L_l$) and spoofing ($S_1, S_2, ..., S_s$) categories, 
			and generates $N$ zero- and few-shot FAS tasks. 
			(b) The meta-learner inner-updates itself on the support set for $u$ steps (the pink arrow), and updates its weight $\theta$ to $\theta^{(u)}$.
			Then we get the meta-learner's zero- and few-shot learning performance and meta-learning loss by testing the updated meta-learner on the query set.
			Finally, we optimize the meta-learner with the meta-learning loss.
			The $L_j$ (Q) in query set of the K-shot FAS task means the query set contains $Q$ faces from the $L_j$ living face category.
		}
		\label{fig:fig2}
	\end{figure*}

	\subsection{Few-shot and Zero-shot Learning}
	\noindent \textbf{Few-shot} learning~\cite{match-network,prototypical}, which aims at learning from very few instances, has attracted lots of attention.
	Metric learning based methods are popular to solve few-shot learning problem.
	These methods train a non-linear mapping function projecting images to an embedding space, and classify the image with nearest neighbor or linear classifier.
	Recently, meta-learning~\cite{On-the-optimiazation,MAML,Reptile,RL2,SNAIL,LLAML,our} based methods solve few-shot learning by training a meta-learner on few-shot learning tasks.
	Given a few examples of new object categories, these methods train the meta-learner to recognize the new categories by memorizing~\cite{SNAIL,RL2} the few examples of the new categories or updating its weight\cite{MAML,Reptile,our}.
	
	\emph{Few-shot learning task} is usually referred to $N$-way $K$-shot learning task, which contains $N$ unseen categories for the model to recognize.
	Compared to conventional classification problem, each way in the task has a relatively smaller number ($K$) of labeled examples provided for training.
	In a nutshell, an $N$-way $K$-shot task provides a support set of $NK$ labeled examples for the model to learn.
	In evaluation, a query set that contains several other examples from the $N$ unseen categories is used to test the model.
	
	\noindent \textbf{Zero-shot} learning which aims at to recognize unseen category with only description or semantic attributes of the new category.
	Similar to metric learning based few-shot learning, traditional zero-shot learning methods train a model to learn a visual-semantic embedding~\cite{zero-DAP,zero-IAP,zero-cnn2}.
	Once the embedding is trained, the instance of unseen classes can be classified in two steps. 
	Firstly, the instance is projected into the semantic space.
	Secondly, it is labeled to the class which has the most similar semantic attributes.
	
	For \emph{Zero-shot learning task},
	the model is required to recognize unseen categories by learning only from the description or semantic information of these unseen categories.
	In other words, the support set of the zero-shot learning task contains only the description or semantic information of these unseen categories.
	In this paper, we prefer to solve both zero- and few-shot FAS problems simultaneously.

	\section{Methodology}
	
	In this section, we detail the proposed Adaptive Inner-update Meta Face Anti-spoofing (AIM-FAS) method.
	
	\subsection{Zero- and Few-shot FAS Problem}
	\noindent \textbf{Zero- and few-shot FAS task} \quad
	We propose that there exist general discriminative features among predefined PAs and unpredicted new PAs.
	In other words, the knowledge in predefined living and spoofing faces can be transferred to detect new living (\emph{e.g.} the living faces recorded in new application scenarios) and new spoofing types. 
	Therefore, we define zero- and few-shot FAS task differently from the traditional zero- and few-shot learning task.
	In zero-shot FAS, the model learns the feature to recognize new living and spoofing categories from predefined living and spoofing categories.
	The support set in zero-shot FAS task only contains predefined living and spoofing faces.
	In few-shot FAS task, the model learns the feature to detect new spoofing types not only from the predefined types but also from a few examples of new living and spoofing types
	The support set in few-shot FAS task contains faces of not only new living and spoofing types but also predefined types.

	\subsubsection{Task generation}
	\label{section:task_generate}
	To generate zero- and few-shot FAS tasks, we split the living and spoofing faces into fine-grained pattern, and show the fine-grained dataset structure in Fig.\ref{fig:fig2}(a).
	We show an example of $K$-shot FAS task in Fig.\ref{fig:fig2}(b), and generate the $K$-shot ($K>=0$) FAS tasks in the following way:
	1) sample one fine-grained living category $L_i$ and one spoofing category $S_m$, from the train set.
	2) sample $M-K$ faces from each of $L_i$ and $S_m$.
	3) re-sample one fine-grained living category $L_j$ and one spoofing category $S_n$.
	Note that, for training tasks, $L_j$ and $S_n$ are sampled from the train set, and for testing tasks, they are sampled from the test set.
	4) sample $K+Q$ faces from each of $L_j$ and $S_n$. 
	5) build the query set with $2Q$ faces from $L_j$ and $S_n$, and build the support set with the other $2 * (M-K) + 2 * K=2M$ faces.
	In other words, $L_i$ and $S_m$ can be seen as the predefined categories, and $L_j$ and $S_n$ can be seen as the new emerged categories.
	In this way, we generate both zero-shot and few-shot learning tasks.
	When $K=0$ (zero-shot FAS), the meta-learner learns from $L_i$ and $S_m$, and predict faces from $L_j$ and $S_n$.
	When $K>0$ (few-shot FAS), the meta-learner learns from $L_i$, $L_j$, $S_m$ and $S_n$, and predict faces from $L_j$ and $S_n$.

	\subsection{AIM-FAS}
	\label{sec:Meta}
	To tackle the zero- and few-shot FAS problem, we develop our Adaptive Inner-update Meta Face Anti-Spoofing (AIM-FAS)
	with training a meta-learner on zero- and few-shot FAS training tasks.
	Furthermore, an Adaptive Inner-Update (AIU) strategy is presented to improve the performance further, as the meta-learner will inner-update more accurately on the support set with AIU.
	Specifically, on a given zero- or few-shot FAS task, one training iteration of the meta-learner consists of two stages.
	
	\noindent \textbf{Inner-update stage} \quad The meta-learner with weight $\theta$ inner-updates itself on the support set for several steps which can be formulated as:
	\begin{equation}
	\mathcal{L}_{s(\tau_i)}(\theta^{(j)}_{i}) \leftarrow \frac{1}{\|s(\tau_i)\|} \sum\nolimits_{x,y \in s(\tau_i)}\emph{l}(\emph{f}_{\theta^{(j)}_{i}}(x),y),
	\label{eq:support_update1}
	\end{equation}
	\begin{equation}
	\theta_i^{(j+1)} \leftarrow\theta_i^{(j)}-\alpha \cdot {\gamma^{j}} \cdot \nabla_{\theta_i^{(j)}}\mathcal{L}_{s(\tau_i)}(\theta_i^{(j)}),
	\label{eq:support_update2}
	\end{equation}
	where $\tau_i$ is a randomly selected zero- or few-shot FAS training task, and $\theta_i^{(j)}$ is the meta-learner's weight after $j$ inner-update steps.
	Note that, for each task $\tau_i$, $\theta_i^{(0)} = \theta$ when $j=0$.
	$x$ and $y$ is a pair of instance and label sampled from the support set of $\tau_i$. 
	$\|s(\tau_i)\|$ is the number of instances of the support set.
	If not otherwise specified, $\|s(\tau_i)\| \!=\!2M$.
	$f_{\theta^{(j)}_{i}}(x)$ is the meta-learner's prediction on instance $x$, and $\mathcal{L}_{s(\tau_i)}(\theta_i^{(j)})$ is the meta-learner's loss on the support set.
	\vspace{3pt}
	
	Scalar parameter $\alpha$ and $\gamma$ in Eq.\ref{eq:support_update2} are the keys to achieve AIU. 
	Both of them are trainable.
	The product of $\alpha$ and $\gamma^{j}$ is the inner-update learning rate (IULR).
	$j$ is the meta-learner's inner-update step.
	The IULR changes along with the updates of $j$.
	For example, when the meta-learner inner-updates itself on the support set for the first step ($j\!\!=\!\!0$), the IULR is $\alpha$ itself.
	When the meta-learner inner-updates itself on the support set for the second step ($j\!\!=\!\!1$), the IULR turns to $\alpha \cdot {\gamma^{1}}$.
	With trainable $\alpha$ and $\gamma$, the meta-learner inner-updates with an adaptive step size.
	After $u$ inner-update steps, the meta-learner update its weight from $\theta$ to $\theta_i^{(u)}$ on the support set with Eq.\ref{eq:support_update1} and Eq.\ref{eq:support_update2}. 
	
	\noindent \textbf{Optimizing stage} \quad The meta-learner is evaluated and optimized on the query set, which contains faces of unseen living and spoofing categories.
	The optimization can be formulated as:
	\begin{equation}
	\mathcal{L}_{q(\tau_i)}(\theta_i^{(u)}) \leftarrow \frac{1}{\|q(\tau_i)\|} \sum\nolimits_{x,y \in q(\tau_i)}^{}\emph{l}(\emph{f}_{\theta_i^{(u)}}(x),y)
	\label{eq:query_update1}
	\end{equation}
	\begin{equation}
	(\theta, \alpha, \gamma) \leftarrow(\theta, \alpha, \gamma) - \beta\cdot\nabla_{(\theta, \alpha, \gamma)}\mathcal{L}_{q(\tau_i)}(\theta_i^{(u)})
	\label{eq:query_update2}
	\end{equation}
	where $x$ and $y$ is a pair of instance and label from the query set of task $\tau_i$. 
	$\|q(\tau_i)\|$ is the number of instances of the query set.
	If not otherwise specified, $\|q(\tau_i)\|$ is $2Q$.
	Note that when the meta-learner is evaluated on the query set, its weight is $\theta_i^{(u)}$, which is updated from $\theta$ with Eq.\ref{eq:support_update2} for $u$ inner-update steps. 
	Further more, in Eq.\ref{eq:query_update2}, $\nabla_{(\theta, \alpha, \gamma)}\mathcal{L}_{q(\tau_i)}(\theta_i^{(u)})$ uses the meta-learner's loss on query to compute the gradient of $\theta$, $\alpha$ and $\gamma$, but not $\theta_i^{(u)}$.
	$\beta$ is the learning rate in the optimizing stage.
	By constantly training the meta-learner on lots of these zero- and few-shot learning tasks, the meta-learner is forced to learn easy fine-tuning weight $\theta$ and propriety $\alpha$ and $\gamma$.
	With weight $\theta$ and the adaptive IULR $\alpha\cdot \gamma^{j}$, the meta-learner updates itself accurately on the support set, and learn the discriminative features to detect unseen spoofing types.
	
	The training process of AIM-FAS is shown in Algorithm~\ref{algorithm:Meta-FAS-BS} and Fig.\ref{fig:fig2}(b).
	We firstly pre-train the meta-learner to learn the prior knowledge about FAS on the train set (line 2 in Algorithm~\ref{algorithm:Meta-FAS-BS}), and secondly meta-train the meta-learner on zero- and few-shot FAS training tasks.
	The testing process of AIM-FAS is shown in Algorithm~\ref{algorithm:Meta-FAS-test}, in which $\emph{P}_{q(\tau_i)}$ is the meta-learner performance on the query set of $\tau_i$.
	$X_{q(\tau_i)}$ and $Y_{q(\tau_i)}$ are the faces and labels in the query set of task $\tau_i$.
	
	\noindent \textbf{Difference between AIM-FAS with the other traditional FAS methods} \quad
	The difference between AIM-FAS with the other traditional FAS is that AIM-FAS trains the meta-learner to focus on learning the discrimination for detecting new spoofing category, from the support set where contains predefined living and spoofing faces and a few or none data of the new living and spoofing categories, while traditional FAS methods train a detector to learn the discrimination for detecting predefined spoofing faces.
	
	\begin{algorithm}[t]
		\caption{AIM-FAS in training stage}
		{\bfseries input:} $K$-shot ($K>=0$) FAS training tasks \emph{$\Psi_t$}, learning rate $\beta$, number of inner-update steps $u$, initial value of AIU parameters $\alpha$ and $\gamma$. \\
		{\bfseries output:} Meta-learner's weight $\theta$, AIU parameters $\alpha$ and $\gamma$. \\
		{\bfseries 1\,\,\,:} initialize $\theta$ and AIU parameters $\alpha$ and $\gamma$. \\
		{\bfseries 2\,\,\,:} pre-train the meta-learner on the train set. \\
		{\bfseries 3\,\,\,:} {\bfseries while not} done {\bfseries do} \\
		{\bfseries 4\,\,\,:}  \ \ \,	  sample batch tasks $\tau$$_i$ $\in$ \emph{$\Psi_t$} \\
		{\bfseries 5\,\,\,:}  \ \ \,   {\bfseries for} each of  $\tau$$_i$  {\bfseries do} \\
		{\bfseries 6\,\,\,:}  \ \ \ \ \,\,  $\theta_i^{(0)} = \theta$ \\
		{\bfseries 7\,\,\,:}  \ \ \ \ \,\,   {\bfseries for} $j < u$  {\bfseries do} \\
		{\bfseries 8\,\,\,:}  \ \ \ \ \,\, \ \ \,	$\mathcal{L}_{s(\tau_i)}(\theta_{i}^{(j)}) \leftarrow \frac{1}{\|s(\tau_i)\|} 
		\sum_{x,y \in s(\tau_i)}^{} \emph{l}(\emph{f}_{\theta_{i}^{(j)}}(x), y) $   \\
		{\bfseries 9\,\,\,:}  \ \ \ \ \,\, \ \ \,  ${\theta_i^{(j+1)}} \leftarrow {\theta_{i}^{(j)}} - \alpha \cdot \gamma^{j} \cdot \nabla_{\theta_i^{(j)}}\mathcal{L}_{s(\tau_i)}({\theta_i^{(j)}})$\\
		{\bfseries 10:}  \ \ \ \ \ \ \,\,\,	$\mathcal{L}_{q(\tau_i )}\!(\theta_i^{(j+1)}) \! \leftarrow  \! \frac{1}{\|q(\tau_i)\|}  \! \sum_{x,y \in q(\tau_i)}^{}  \! \emph{l}(\emph{f}_{\theta_i^{(j\!+\!1)}}\!(x),y) $\\
		{\bfseries 11:}  \ \ \ \ \,\, \ \ \, $j = j + 1 $\\
		{\bfseries 12:} 	\ \ \ \ \,\, {\bfseries end} \\
		{\bfseries 13:} 	\ \ \, {\bfseries end} \\
		{\bfseries 14:} 	\ \ \, $(\theta, \alpha, \gamma) \leftarrow (\theta, \alpha, \gamma)$ - $\beta \cdot \nabla_{(\theta, \alpha, \gamma)}\sum_{\tau_i}^{}\mathcal{L}_{q(\tau_i)}(\theta_i^{(u)}$)\\
		{\bfseries 15:} 	{\bfseries end}
		\label{algorithm:Meta-FAS-BS}
	\end{algorithm}

	\begin{algorithm}[!t]
		\caption{AIM-FAS in testing stage}
		{\bfseries input:} $K$-shot FAS testing tasks \emph{$\Psi_v$}, number of inner-update steps $u$, Meta-learner's weight $\theta$, AIU parameters $\alpha$ and $\gamma$. \\
		{\bfseries output:} Meta-learner's performance \emph{P}. \\
		{\bfseries 1\,\,\,:}   {\bfseries for} each of  $\tau$$_i$ $\in$ \emph{$\Psi_v$} {\bfseries do} \\
		{\bfseries 2\,\,\,:}  \ \ \,  $\theta_i^{(0)} = \theta$ \\
		{\bfseries 3\,\,\,:}  \ \ \,   {\bfseries for} $j < u$  {\bfseries do} \\
		{\bfseries 4\,\,\,:}  \ \ \ \ \,\, 	$\mathcal{L}_{s(\tau_i)}(\theta_{i}^{(j)}) \leftarrow \frac{1}{\|s(\tau_i)\|} 
		\sum_{x,y \in s(\tau_i)}^{} \emph{l}(\emph{f}_{\theta_{i}^{(j)}}(x), y) $   \\
		{\bfseries 5\,\,\,:}  \ \ \ \ \,\,   ${\theta_i^{(j+1)}} \leftarrow {\theta_{i}^{(j)}} - \alpha \cdot \gamma^{j} \cdot \nabla_{\theta_i^{(j)}}\mathcal{L}_{s(\tau_i)}({\theta_i^{(j)}})$\\
		{\bfseries 6\,\,\,:}    \ \ \ \ \,\,  $j = j + 1 $\\
		{\bfseries 7\,\,\,:} 	\ \ \, {\bfseries end} \\
		{\bfseries 8\,\,\,:} 	\ \ \, \emph{P}$_{q(\tau_i)} \leftarrow \emph{p}(\emph{f}_{\theta_i^{(u)}}(X_{q(\tau_i)}),Y_{q(\tau_i)}) $\\
		{\bfseries 9\,\,\,:} 	 {\bfseries end} \\
		{\bfseries 10:} 	 $P \leftarrow \frac{1}{\|\Psi_v\|} \sum_{\tau_i \in \Psi_v}^{}$\emph{P$_{q(\tau_i)}$}
		\label{algorithm:Meta-FAS-test}
	\end{algorithm}

	\subsubsection{Fusion Train (FT)}
	Traditionally, meta-learning methods train meta-learners independently for different $K$-shot ($K>0$) learning problems. 
	For example, to solve 1-shot learning problem, they usually train a meta-learner on 1-shot training tasks, and to solve 5-shot learning problem, they train another meta-learner on 5-shot training tasks.
	In contrast, our goal is training one meta-learner to solve both zero- and few-shot FAS tasks.
	So, in AIM-FAS, we train the meta-learner in a Fusion Training (FT) manner, which means the meta-learner is simultaneously trained on different $K$-shot ($K>=0$) FAS tasks.
	Specifically, the meta-learner is trained on tasks of both zero- and few-shot FAS tasks, \emph{ie.} 0-shot, 1-shot, 2-shot, \emph{etc.}.
	In our experiment, we show that FT improves AIM-FAS on both zero- and few-shot FAS tasks.
	
	\subsubsection{Network}
	
\begin{figure}[t]
		\centering    
		\includegraphics[width=0.95\columnwidth]{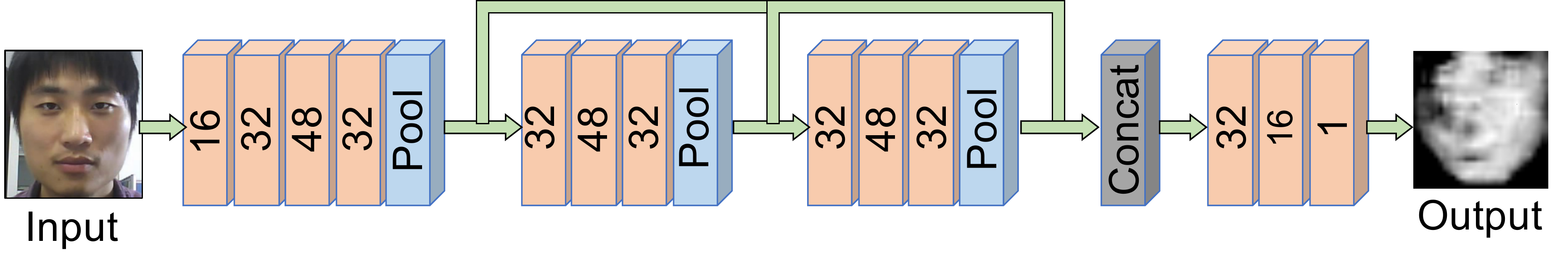}
		\caption{Network structure of AIM-FAS. 
			The pink cube is the convolution layer, on which the number means the number of channels of its filter.} 
		\label{fig:network}  
	\end{figure}

	Depth-supervised FAS methods~\cite{Liu2018Learning} take advantage of the discrimination between spoofing and living faces based on 3D shape, and provide more detailed information for the FAS model to capture spoofing cues. 
	Motivated by this, AIM-FAS trains the meta-learner to solve depth regression based zero- and few-shot FAS tasks.
	We build a depth regression network for AIM-FAS and name it as FAS-DR.
	The structure and detail of FAS-DR is shown in Fig.\ref{fig:network}.
	There are three cascaded blocks in the network backbone, and all their features are concatenated for predicting the facial depth. 
	We formulate the facial depth prediction process as \emph{$\widetilde{D}$} = $f_{\theta}(x)$, where \emph{$x$} $\in$ $\mathbb{R}^ {256 \times 256 \times 3}$ is the RGB facial image, and \emph{$\widetilde{D}$} $\in \mathbb{R} ^ {32 \times 32 \times 1}$ is the predicted facial depth, and $\theta$ is the network's weights.
	
	Contrastive Depth Loss (CDL)~\cite{Wang2018Exploiting} is utilized to help the meta-leaner to predict vivid facial depth.
	The CDL is 
	\begin{equation}
	L^{contrast} = \sum_{i}^{}\|k_i^{contrast}\cdot \widetilde{D} - k_i^{contrast}\cdot D\|_2^2,
	\label{eq:contrastive}
	\end{equation}
	where $D$ is the generated ``ground truth" facial depth label. \emph{$k_i^{contrast}$} is the kernel of CDL, and i $\in$ \{0,1,2,3,4,5,6,7\}.

	\section{Zero- and Few-shot FAS Benchmarks}
	\label{benchmark}
	To verify AIM-FAS, we propose three \textbf{Z}ero- and \textbf{F}ew-shot FAS benchmarks:\emph{OULU-ZF}, \emph{Cross-ZF} and \emph{SURF-ZF}.
	
	\textbf{\emph{OULU-ZF}} is a single domain zero- and few-shot FAS benchmark and is build based on OULU-NPU.
	In OULU-NPU, there are 6 image capture devices, 3 kinds of living faces (living faces captured within 3 different sessions), 2 kinds of print attacks, and 2 kinds of replay attacks.
	All living and spoofing faces are captured with 55 people.
	We reorganize OULU-NPU into \emph{OULU-ZF} and show the structure of \emph{OULU-ZF} in Tab.\ref{tab:single-domain}.
	There is no overlap between the train (seen categories) and test (unseen categories) set.
	The train set contains 2 kinds of living face (living 2 and 3), 1 kind of print faces and replay faces.
	All living and spoofing faces in the train set are captured with device 1,2,4,5 and 6.
	Whereas, in the test set, all living faces are the living 1 category.
	
	\textbf{\emph{Cross-ZF}} is a cross domain zero- and few-shot FAS benchmark which is more challenging than \emph{OULU-ZF}.
	It contains more varied living and spoofing categories.
	We build \emph{Cross-ZF} based on several public FAS dataset.
	Tab.\ref{tab:single-domain} shows the structure of \emph{Cross-ZF}.
	The train set contains 7 kinds of living faces, 4 kinds of printed faces, and 7 kinds of replayed faces, from three public dataset: CASIA-MFSD, MSU-MFSD, and SiW.
	The test set contains living and spoofing faces from the other three dataset: 3DMAD, Oulu-NPU, and Replay-Attack.
	There is no overlap between the train set and test set, and the test set contains 3D Mask faces, which are different greatly with printed and replayed faces.
	
	\textbf{\emph{SURF-ZF}} is a cross modal zero- and few-shot FAS benchmark.
	We build \emph{SURF-ZF} based on the CASIA-SURF dataset.
	Structure of \emph{SURF-ZF} is shown in Tab.\ref{tab:cross-modal}.
	We extract several samples from CASIA-SURF, and split these examples into train, validation, and test set.
	The train set contains RGB and Depth modalities, and the test/validation set contains IR and depth modalities.
	Each set contains all PSAIs (Living 1;Print1;Cut1-5).
	Based on {\em SURF-ZF}, we can test the model's ability of learning fast from new modalities.

	\section{Experiments}
	\subsection{Experiment Setup}
	\noindent \textbf{Performance Metrics}
	In our experiments, AIM-FAS is evaluated by: 
	1) Attack Presentation Classification Error Rate ($APCER$);
	2) Bona Fide Presentation Classification Error Rate ($BPCER$);
	3) $ACER$~\cite{ACER}, which evaluates the mean of $APCER$ and $BPCER$.
	4) Area Under Curve (AUC).
	\noindent \textbf{Evaluation Process}  \quad
	On all benchmarks, we evaluate the meta-learner's zero- and few-shot FAS performance in the following way:
	1) train the meta-learner on the training tasks generated on the train set;
	2) test the meta-learner on zero- and few-shot FAS testing tasks on the test set;
	3) calculate the meta-learner's performance with Eq.\ref{eq:acer_final}.
	\begin{equation}
	\begin{split}
	{ACER}_{avg} \quad = \quad \sum{{ACER}}^T_{i=1}/T, \\
	ACER = ACER_{avg} \pm 1.96 * \sigma/\sqrt{T}
	\end{split}
	\label{eq:acer_final}
	\end{equation}
	
	$\sigma$ is the standard deviation of \emph{$ACER$} on all the test tasks, and $T$ is the quatity of test tasks. 
	
	\begin{table}[t]
		\centering
		\caption{Zero- and few-shot FAS benchmark: {\em OULU-ZF}.}	
		\resizebox{0.9\columnwidth}{!}{
			\begin{tabular}{|c|c|c|c|}
				\hline
				Set & Device & Subjects &$\qquad\qquad$PSAI$\qquad\qquad$  \\
				\hline 
				Train &Phone 1,2,4,5,6 &1-20 &Living 2,3; Print 1;Replay 1  \\
				\hline
				Val &Phone 3 &21-35 &Living 1-3; Print 1,2; Replay 1,2 \\
				\hline
				Test &Phone 1,2,4,5,6 &36-55 &Living 1; Print 2; Replay 2\\
				\hline
			\end{tabular}
		}
		\label{tab:single-domain}
	\end{table}

	\begin{table}[t]
		\centering
		\caption{Zero- and few-shot FAS benchmark: {\em Cross-ZF}.}	
		\resizebox{0.95\columnwidth}{!}{
			\begin{tabular}{|c|c|c|}
				\hline
				Set & Domains &$\qquad\qquad$PSAI$\qquad\qquad$  \\
				\hline 
				Train &CASIA-MFSD, MSU-MFSD, SiW  &Living 1-7; Print 1-4; Replay 1-7  \\
				\hline
				Val &MSU-USSA &Living 1;Print 1-2; Replay 1-6 \\
				\hline
				Test &3DMAD, Oulu-NPU, Replay-Attack &Living 1-9; 3D Mask; Print 1-3;Replay 1-4\\
				\hline
			\end{tabular}
		}
		\label{tab:cross-domain}
	\end{table}

	\begin{table}[!t]
		\centering
		\caption{Zero- and few-shot FAS benchmark: {\em SURF-ZF}.}
		\resizebox{0.63\columnwidth}{!}{
			\begin{tabular}{|c|c|c|}
				\hline
				Set &Modals &PSAI  \\
				\hline
				Train &RGB;Depth &Living 1;Print 1;Cut 1-5  \\
				\hline
				Val &IR;Depth &Living 1;Print 1;Cut 1-5  \\
				\hline
				Test &IR:Depth &Living 1;Print 1;Cut 1-5 \\
				\hline
			\end{tabular}
		}
		\label{tab:cross-modal}
	\end{table}

\begin{figure}[t]
		\centering
		\includegraphics[width=0.8\columnwidth]{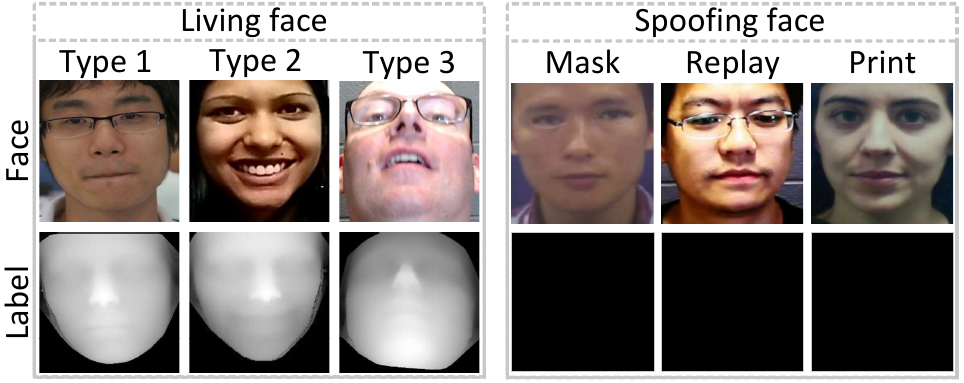}
		\caption{Generated depth label of living and spoofing faces.
		}
		\label{fig:face_depth}
	\end{figure}

	\noindent \textbf{Implementation Details} \quad
	In our experiment, we generate the ground-truth depth label of living face with the PRNet~\cite{Feng2018Joint}, and normalize the generated facial depth to [0, 1].
	To distinguish spoofing face from living face, we set the ground-truth depth label of spoofing face to all zero.
	The generated facial depths are shown in Fig.~\ref{fig:face_depth}.
	All the facial depth maps are resized into 32${\times}$32 resolution.
	
	We generate 100,000 training tasks on the train set and 100 ($T=100$) testing tasks on the test set.
	For each $K$-shot training task, $K$ is randomly sampled from \{0,1,3,5,7,9\}.
	For testing tasks, $K$ is a specified number indicating the meta-learner is tested on specified $K$-shot tasks.
	For example, if we evaluate the meta-learner's performance on zero-shot FAS tasks, we set $K=0$ and generate 100 such zero-shot testing tasks to test the meta-learner.
	We set $Q$ to 15, $M$ to 10.
	The meta batch size is set to 8, and the meta-learning rate $\beta$ is set to 0.0001.
	The AIU parameters $\alpha$ and $\gamma$ are initialized to 0.001 and 1, respectively.
	
	\noindent \textbf{Compared Methods} \quad
	To validate the performance of AIM-FAS on zero- and few-shot FAS problem, we compare AIM-FAS with three FAS detectors Resnet-10, FAS-DR, and DTN*.
	The detector FAS-DR is the network of AIM-FAS trained in traditional supervised learning.
	As the network of detector FAS-DR is the same as that of AIM-FAS, We treat detector FAS-DR as the baseline of AIM-FAS.
	The detector Resnet-10 is a binary classification FAS model and is also trained in traditional supervised learning.
	DTN\cite{DTN} is a zero-shot FAS detector.
	We re-implement DTN with all experiment settings the same to the original paper and named it as DTN*.
	For fairly comparison, we set up the evaluation protocol for all methods, which is shown in Tab.\ref{tab:compared methods}.
	For example, the detector Resnet-10 is trained on the train set, and to evaluate its 0-shot performance, we evaluate it directly on the query set of 0-shot FAS tasks without finetuning on the support set.
	To evaluate its 1-shot performance, we first finetune it on the support set of the 1-shot tasks and then evaluate it on the corresponding query set.
	
	\begin{table}
		\centering
		\caption{Evaluation detail of compared methods and AIM-FAS.}
		\resizebox{0.98\columnwidth}{!}{
			\begin{tabular}{|c|c|c|c|c|c|}
				\hline
				\multirow{2}{*}{Method} & \multirow{2}{*}{Train} & \multicolumn{2}{c|}{0-shot Test} & \multicolumn{2}{c|}{1- or 5-shot Test} \\
				\cline{3-6}
				& & Finetune & Evaluate & Finetune & Evaluate \\
				\hline
				Compared & Train set & $/$ & Query & Support & Query \\
				\cline{1-6}
				\textbf{AIM-FAS} & Training tasks  & Support & Query & Support & Query \\
				\hline
			\end{tabular}
		}
		\label{tab:compared methods}
	\end{table}

	\subsection{Experiment on Proposed Benchmarks}
	Corresponding experimental results on the proposed benchmarks are shown in Tab.\ref{tab:cross-domain result}.
	It can be seen that AIM-FAS outperforms the other detectors with a clear margin on all benchmarks. 
	Note that, as original DTN is designed for zero-shot FAS, we follow the same way to evaluate DTN* on zero-shot instead of few-shot tasks.
	Compared with FAS-DR, the {\em ACER} of AIM-FAS decreases by 25\%, 17\%, and 28\% on zero-, 1- and 5-shot tasks on \emph{OULU-ZF}, respectively, 
	and decreases by 38\%, 30\%, and 38\% on \emph{Cross-ZF},
	and decreases by 12\%, 13\%, and 16\% on \emph{SURF-ZF}. 
	Since AIM-FAS trains the meta-learner to focus on learning discrimination of new spoofing types from predefined faces, or from predefined faces and a few examples of new living and spoofing types.
	It learns more generalized discriminative features for detecting new attack types.
	Whereas, FAS-DR only focus on learning the discrimination to distinguish predefined spoofing faces from living faces.
	
	Another phenomenon is that the margin between AIM-FAS and the other methods is more clear on \emph{Cross-ZF} than on the other benchmarks.
	The possible reason behind is that \emph{Cross-ZF} contains more diverse fine-grained living and spoofing categories, which is more suitable than the other benchmarks for AIM-FAS learning general discrimination for detecting new attack types.
	
	\begin{table}
		\centering
		\caption{Experimental result on three proposed benchmarks. DTN* is our re-implementation of DTN\cite{DTN} with the same setting as its original paper.
		}
		\resizebox{0.98\columnwidth}{!}{
			\begin{tabular}{|c|c|c|c|c|}
				\hline
				\multirow{2}{*}{Benchmark} &\multirow{2}{*}{Method} &\multicolumn{3}{c|}{ACER(\%)} \\
				\cline{3-5}
				& &0-shot &1-shot &5-shot \\
				\hline
				\multirow{4}{*}{OULU-ZF} &Resnet-10 & 7.27$\pm$1.42 & 7.13$\pm$1.19 & 4.20$\pm$1.12 \\
				\cline{2-5}
				& DTN* & 5.83$\pm$1.16\ & $/$ & $/$ \\
				\cline{2-5}
				& FAS-DR & 6.60$\pm$1.78\ & 4.83$\pm$1.40\ & 3.37$\pm$1.02\ \\
				\cline{2-5}
				&\textbf{AIM-FAS} & \textbf{4.97$\pm$1.29}  &\textbf{4.00$\pm$1.31} &\textbf{2.44$\pm$0.71} \\
				\cline{2-5}
				\hline
				\multirow{4}{*}{Cross-ZF} &Resnet-10 & 26.51$\pm$2.59 & 26.37$\pm$2.64 & 15.43$\pm$2.38 \\
				\cline{2-5}
				& DTN* & 17.62$\pm$3.89 & $/$ & $/$ \\
				\cline{2-5}
				& FAS-DR & 13.49$\pm$3.62 & 10.54$\pm$3.03 & 7.21$\pm$1.87 \\
				\cline{2-5}
				&\textbf{AIM-FAS} & \textbf{8.43$\pm$2.92}  &\textbf{7.34$\pm$1.45} &\textbf{3.11$\pm$1.01} \\
				\cline{2-5}
				\hline   
				\multirow{4}{*}{SURF-ZF} &Resnet-10 & 45.60$\pm$0.72  & 45.43$\pm$0.81 & 44.53$\pm$1.40\\
				\cline{2-5}
				& DTN* & 45.27$\pm$2.29\ & $/$ & $/$ \\
				\cline{2-5}
				& FAS-DR & 34.61$\pm$2.15 & 33.17$\pm$2.07 & 32.50$\pm$1.51 \\
				\cline{2-5}
				&\textbf{AIM-FAS} & \textbf{30.97 $\pm$1.28}  & \textbf{28.75 $\pm$1.49} & \textbf{27.27$\pm$1.25} \\
				\cline{2-5}
				\hline
			\end{tabular}
		}
		\label{tab:cross-domain result}
	\end{table}

	\subsection{Experiment on Existing Dataset}
	To further evaluate the advantages of AIM-FAS, we test AIM-FAS on the protocol proposed by \cite{zero-protocol-FAS}.
	In this protocol, CASIA, Replay-Attack, and MSU-MFSD are used to evaluate the FAS model's zero-shot performance across replay and print attacks. 
	As shown in Tab.\ref{tab:comparison with DTN}, AIM-FAS performs better than the other methods on most sub-protocols with rising the average AUC by at least 0.52\%.
	The result further reveals that AIM-FAS is successful for not only few-shot but also zero-shot FAS.
	AIM-FAS performs not the best on the sub-protocols of CASIA Video, Replay-Attack Video, and MSU Printed Photo. 
	The possible reason is that the training spoofing categories of these sub-protocols are unitary and not suitable for AIM-FAS learning the discrimination to detect the testing spoofing category.
	For example, on CASIA, the Cut Photo and Warped Photo spoofing categories are not varied enough, and the meta-learner trained on these categories can hardly summarize and capture the general discrimination that is effective for detecting the Video category.
	
	\begin{table*}
		\centering
		\caption{Performance of AIM-FAS on CASIA, Replay-Attack, and MSU-MFSD. 
			The evaluation metric is AUC(\%).
		}
		\resizebox{2.05\columnwidth}{!}{
			\begin{tabular}{|c|c|c|c|c|c|c|c|c|c|c|}
				\hline
				\multirow{2}{*}{Methods} &\multicolumn{3}{c|}{CASIA} &\multicolumn{3}{c|}{Replay-Attack} &\multicolumn{3}{c|}{MSU} & \multirow{2}{*}{Overall} \\
				\cline{2-10}
				&Video &Cut Photo &Warped Photo &Video &Digital Photo &Printed Photo &Printed Photo &HR Video &Mobile Video & \\
				\hline
				OC-SVM\_{RBF}$+$BSIF  & 70.7  & 60.7 & 95.9 & 84.3 & 88.1 & 73.7 & 64.8 & 87.4 & 74.7 & 78.7$\pm$11.7 \\
				\hline
				SVM\_{RBF}$+$BSIF & 91.5 & 91.7 & 84.5 & 99.1 & 98.2 & 87.3 & 47.7 & 99.5 & 97.6 & 88.6$\pm$16.3  \\
				\hline
				NN+LBP  & \textbf{94.2} & 88.4 & 79.9 & 99.8 & 95.2 & 78.9 & 50.6 & 99.9 & 93.5 & 86.7$\pm$15.6 \\
				\hline
				DTN & 90.0 & 97.3 & 97.5 & \textbf{99.9} & 99.9 & 99.6 & \textbf{81.6} & 99.9 & 97.5 & 95.9$\pm$6.2 \\
				\hline
				\textbf{AIM-FAS(ours)} & 93.6  & \textbf{99.7} & \textbf{99.1} & 99.8 & \textbf{99.9} & \textbf{99.8} & 76.3 & \textbf{99.9} & \textbf{99.1} & \textbf{96.4$\pm$7.8} \\
				\hline
			\end{tabular}
		}
		\label{tab:comparison with DTN}
	\end{table*}
	
	\subsection{Ablation Study}
	\noindent \textbf{AIM-FAS for Binary Supervision} \quad
	We validate the effectiveness of AIM-FAS on binary supervised architecture by taking Resnet-10 as the backbone, named as AIM-FAS (Resnet). And Resnet-10 trained in traditional supervised manner is set as the baseline of AIM-FAS (Resnet). The comparison of these two methods on \emph{Cross-ZF} is shown in Tab.\ref{tab:ablation on Cross-ZF}. Compared with Resnet-10, AIM-FAS (Resnet) decrease the ACER by 45.08\%, 54.72\%, and 41.55\% on 0-shot, 1-shot, and 5-shot tasks, respectively. This demonstrates the generality of AIM-FAS on different network structures and different supervision manners. 
	
	\noindent \textbf{Effectiveness of Predefined Living and Spoofing Faces in Support Set} \quad
	Here we verify whether predefined living and spoofing faces in the support set are useful for AIM-FAS to learn discrimination to detect the new spoofing category.
	In this experiment, during the testing stage, we generate $K$-shot FAS tasks without predefined living and spoofing categories in the support set.
	In other words, the support set of $K$-shot FAS tasks here contains no categories $L_i$ and $S_m$ in Fig.\ref{fig:fig2}(b).
	For $K$-shot ($K>0$) FAS tasks, AIM-FAS without PreDefined living and spoofing faces (named as AIM-FAS w/o PD) inner-updates the meta-learner with only $2K$ new type of spoofing/living faces, and then test the meta-learner on the query set. 
	Note that we do not test AIM-FAS w/o PD on zero-shot FAS tasks since that support set of zero-shot FAS tasks here is empty. 
	In Tab.\ref{tab:ablation on Cross-ZF}, we can see that AIM-FAS w/o PD increases the ACER(\%) by 11\% and 42\% on 1-shot and 5-shot FAS, respectively. 
	The worse performance of AIM-FAS w/o PD indicates that the predefined living and spoofing faces indeed bring benefit for the meta-learner learning discrimination for detecting new attacks.
	
	\noindent \textbf{Effectiveness of Fusion Training (FT)} \quad
	For the trained meta-learner to be capable of solving both zero- and few-shot FAS problems, we present an FT strategy in AIM-FAS for training the meta-learner simultaneously on all $K$-shot tasks.
	To assess FT, we conduct an experiment that trains the meta-learner without FT.
	AIM-FAS w/o FT trains a meta-learner on 0-shot tasks for 0-shot testing, and train another meta-learner on 1-shot tasks for 1-shot testing, and so on. 
	Tab.\ref{tab:ablation on Cross-ZF} shows the performance of AIM-FAS w/o FT.
	Compared with AIM-FAS w/o FT, the AIM-FAS performs better on all kinds of shot scenes. 
	The possible reason is that the FT manner provides the meta-learner diverse $K$-shot FAS tasks so that the AIM-FAS meta-learner generalizes better from training tasks to testing tasks.
	In other words, with FT, the testing shot scene is a subset of the training shot scenes.

\begin{figure}[t]
		\centering
		\includegraphics[width=0.8\columnwidth]{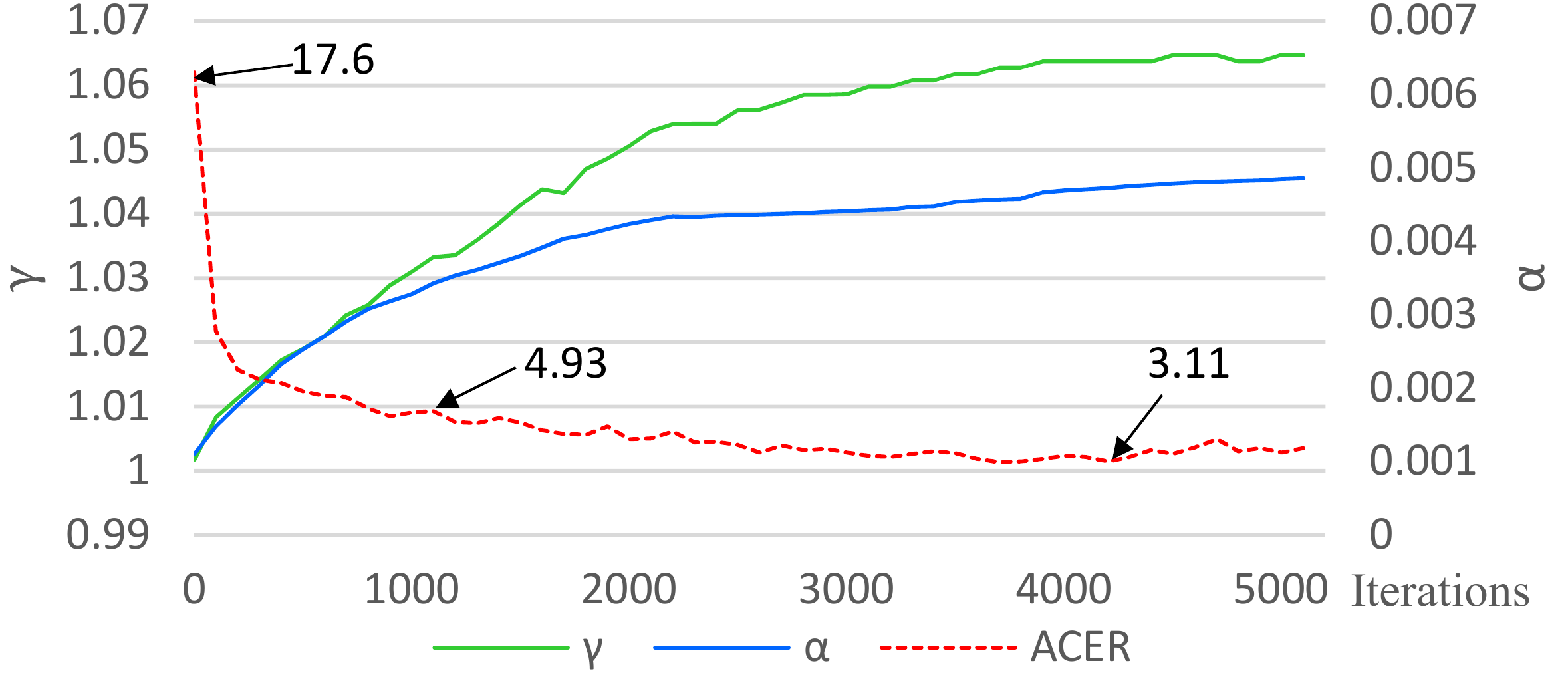}
		\caption{
			The learning curve of $\alpha$, $\gamma$ and the meta-learner's ACER.
			The left Y-axis is the Y-axis of $\gamma$.
			The right Y-axis is the Y-axis of $\alpha$.
			The X-axis is the training iteration.
		}
		\label{fig:gamma_alpha}
	\end{figure}

	\noindent \textbf{Impact of Adaptive Inner-Update (AIU)} \quad
	In this experiment, we discard the Adaptive Inner-Update (AIU) from the complete AIM-FAS. 
	Tab.\ref{tab:ablation on Cross-ZF} shows the comparison of AIM-FAS and AIM-FAS w/o AIU. We find that the AIU improves our AIM-FAS with a large margin. 
	Specifically, AIM-FAS with AIU apparently decreases ACER(\%) more than 3.0 on 0-, 1- and 5-shot. Furthermore, in Fig.\ref{fig:gamma_alpha}, we show the curves of $\alpha$ and $\gamma$ of Eq.~\ref{eq:support_update2} during meta-training process. Both $\alpha$ and $\gamma$ present a rising tendency, meanwhile the ACER falls down.
	This indicates that AIM-FAS prefers a larger inner-update learning rate, and with the learned $\alpha$ and $\gamma$, AIM-FAS performs better than the AIM-FAS w/o AIU.

	\begin{table}
		\centering
		\caption{Ablation experiment on \emph{Cross-ZF}. }
		\resizebox{0.95\columnwidth}{!}{
			\begin{tabular}{|l|c|c|c|}
				\hline
				\multirow{2}{*}{Method}  &\multicolumn{3}{c|}{ACER (\%)} \\
				\cline{2-4} &0-shot &1-shot &5-shot \\
				\hline
				Resnet-10 	&  26.51$\pm$2.59 & 26.37$\pm$2.64 & 15.43$\pm$2.38 \\
				\hline
				AIM-FAS(Resnet) & {\bfseries 14.56$\pm$2.63}	&{\bfseries 11.94$\pm$1.85}	&{\bfseries 9.02$\pm$1.69} \\
				\hline
				AIM-FAS w/o AIU 	     &   11.67$\pm$2.08      &   11.53$\pm$2.96     & 6.44$\pm$1.57 \\
				\hline
				AIM-FAS w/o FT & 12.61$\pm$1.67  &9.15$\pm$1.73 &3.25$\pm$1.05 \\
				\hline
				AIM-FAS w/o PD & $/$  & 8.25$\pm$1.16 &  7.66$\pm$2.45   \\
				\hline
				\textbf{AIM-FAS} & \textbf{8.43$\pm$2.92}  &\textbf{7.34$\pm$1.45} &\textbf{3.11$\pm$1.01} \\
				\hline
			\end{tabular}
		}
		\label{tab:ablation on Cross-ZF}
	\end{table}

	\subsection{Visualization and Analysis}
	
\begin{figure}[t]
		\centering
		\includegraphics[width=0.9\columnwidth]{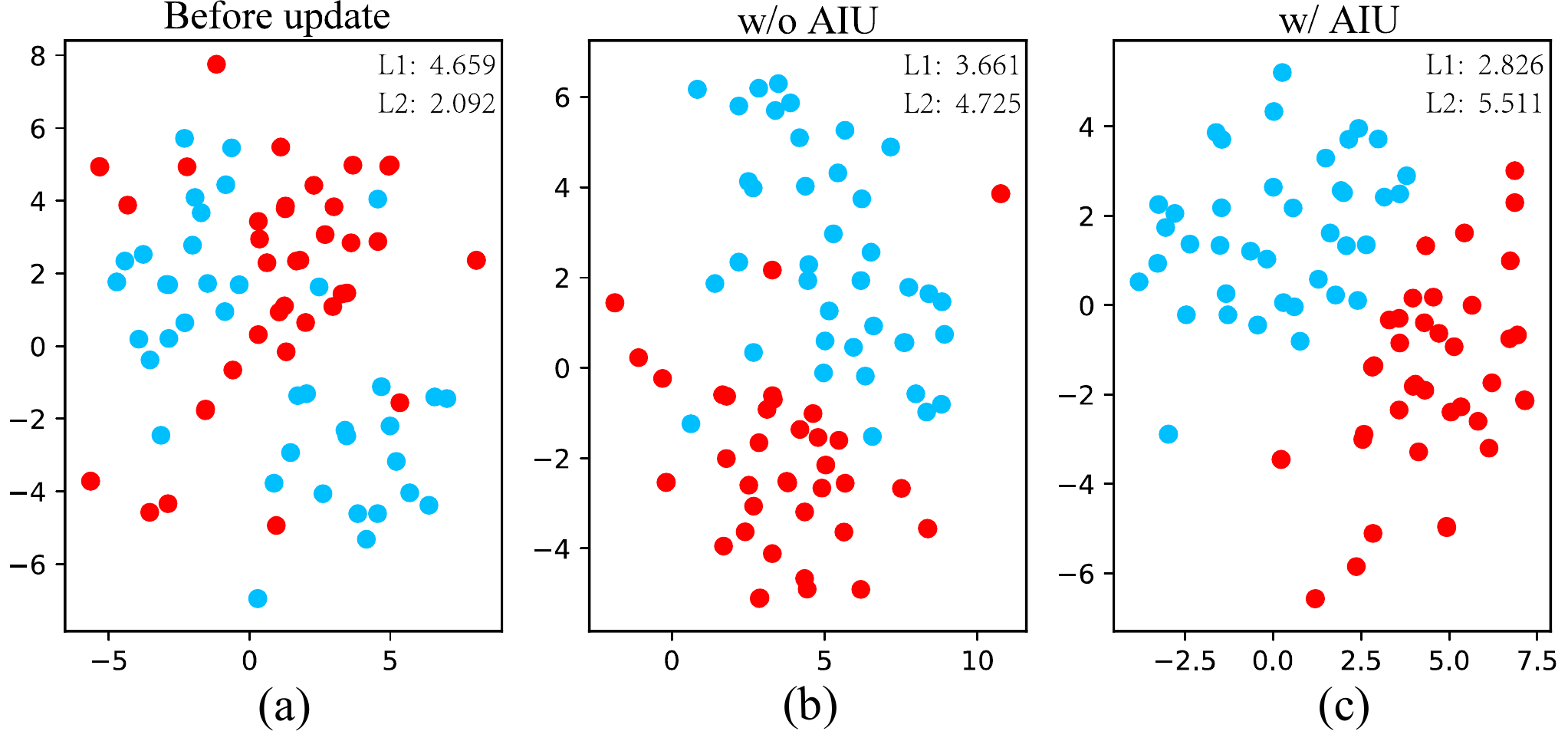}
		\caption{
			Visualization of the distribution of living and spoofing faces in the query set of a 5-shot FAS testing task.
			Color used: \emph{red}=living, \emph{blue}=spoofing.
		}
		\label{fig:cluster}
	\end{figure}

	In this subsection, the feature (feature of the last but one layer) distribution of the meta-learner is illustrated in Fig.\ref{fig:cluster}. 
	We randomly generate a 5-shot FAS testing task and update the meta-learner for 50 inner-update steps on the support set.
	Then the feature distribution of the query set is visualized with t-SNE\cite{tSNE}.
	Fig.\ref{fig:cluster}a is the feature distribution of the query set before the meta-learner updates itself on the support set.
	Fig.\ref{fig:cluster}b and Fig.\ref{fig:cluster}c are the distributions after the meta-learner updates itself for 50 inner-update steps without and with AIU, respectively.
	We also show the category inner distance (L1) and the inter distance (L2).
	From left to right, the distinction between the distribution of living and spoofing faces turns more and more clear, and L1 declines gradually, whereas L2 rises.
	The visualization clearly reveals that the meta-learner learns the discrimination between new living and spoofing categories on the support set, and the AIU helps the meta-learner to learn better discrimination.
	
	\section{Conclusion and Future Work}
	In this paper, we redefine the face anti-spoofing (FAS) as a simultaneously zero- and few-shot learning issue.
	To address this issue, we develop a novel method Adaptive Inner-update Meta Face Anti-Spoofing (AIM-FAS) and propose three zero- and few-shot FAS benchmarks.
	To validate AIM-FAS, we conduct experiments on both the proposed benchmarks and existing zero-shot protocols.
	All experiments show that AIM-FAS outperforms existing methods with a clear margin on both zero- and few-shot FAS.
	In the future, we will develop AIM-FAS to more challenging and practical application scenes.

	\section{Acknowledgments}
	This work was supported by the Chinese National Natural Science Foundation Projects (Grant No. 61876178, 61872367, and 61806196).
	
	\bibliographystyle{aaai}
	\bibliography{main}
	
\end{document}